# Maximum Likelihood Bounded Tree-Width Markov Networks

Nathan Srebro
Laboratory for Computer Science
Massachusetts Institute of Technology
Cambridge, MA 02139

## Abstract

We study the problem of projecting a distribution onto (or finding a maximum likelihood distribution among) Markov networks of bounded tree-width. By casting it as the combinatorial optimization problem of finding a maximum weight hypertree, we prove that it is NP-hard to solve exactly and provide an approximation algorithm with a provable performance guarantee.

## 1 Introduction

Estimating model structure from data remains a difficult problem. A typical approach involves heuristic search using some structure scoring metric, such as the Bayesian score. While this general problem is known to be hard [Chi96] (and the typical approach reflects this fact), hardness of many restricted learning scenarios is not well-understood, nor is the potential for guaranteed approximation algorithms. We focus here on a slightly simpler problem, in which regularization is attained solely by limiting the model to a restricted "concept" class and a maximum likelihood model is sought. Casting the associated estimation problem in purely combinatorial terms permits us to obtain both hardness results and provable guarantees of approximation accuracy.

In 1968, Chow and Liu [CL68] provided a rigorous analysis for finding maximum likelihood Markov trees, casting it as a problem of finding a maximum-weight tree, and thus providing an efficient and exact algorithm for it. We would like to generalize the work of Chow and Liu to the problem of learning a maximum likelihood Markov network of bounded complexity. In Section 2 we discuss how *tree-width* is in many ways the appropriate measure of complexity. Accordingly, we analyze the problem of learning a maximum likelihood Markov network of bounded tree-width.

In fact, we study a somewhat more general problem. The maximum likelihood distribution is a distribution minimizing the information divergence from the empirical distribution. Finding a maximum likelihood distribution can thus be seen as a special case of the problem of *projecting* a target distribution onto a concept class, i.e. finding the distribution from within the class that minimizes the information divergence from the target. Such projections have applications beyond finding the maximum likelihood distribution. In this paper, we use this framework and discuss the problem of projecting a target distribution onto the class of Markov networks of bounded tree-width.

Similarly to the work of Chow and Liu, we are able to formalize the projection problem as a combinatorial optimization problem on graphs. We show that projecting a distribution onto Markov networks of bounded tree-width is equivalent to finding a maximum-weight *hypertree*. This equivalence gives rise to global, integer programming-based approximation algorithms with provable performance guarantees. This contrasts with previously suggested local-search heuristics for the same problem [Mal91]. The equivalence also allows us to study the computational hardness of the learning problem. We show that learning a maximum likelihood Markov network of bounded tree-width is NP-hard, even for tree-width two.

Several other extensions to the work of Chow and Liu have been proposed. Meila [MP99] suggested modeling distributions as mixtures of tree-shaped Markov networks. Dasgupta [Das99] suggested poly-tree Bayesian networks (trees with oriented edges), proving the hardness of this problem.

Complementary to this presentation is a paper on approximation algorithms for the maximum-weight hypertree problem [KS01], which details the algorithms mentioned here. The algorithms are motivated by the problems discussed here, and some results from Section 3 are quoted in [KS01].

It should be noted that this work is concerned with finding distributions, and not with reconstructing the "true" structure. Maximum likelihood is generally not appropriate for model selection, as it will always find maximal structures, even when there is no support for some of the edges. In this sense, too, this work is an extension of Chow and Liu's approach.

## 2 Bounding the Complexity of Markov Networks

A Markov network over a specified graph $G$ is determined by the marginal distributions over cliques in $G$, and this representation essentially gives the number of parameters of Markov networks over $G$. This number is exponential in the clique sizes, and so we would like to keep the clique sizes small.

But other than bounding the number of parameters, we would also like to limit ourselves to tractable computations. Although the clique marginals provide a compact representation for any Markov network with small clique size, there is no generally efficient way of performing exact computations (e.g. of marginal or conditional probabilities) on such graphs. In fact, even calculating the minimum information divergence to Markov networks over the graph (i.e. the maximum likelihood, if the target is the empirical distribution) might be infeasible. Though theoretically possible, it would be extremely optimistic to hope that finding the graph that minimizes this quantity would be easier than calculating it.

In order to work with Markov networks, and in particular to calculate marginal and conditional probabilities, one usually triangulates the graph. On a triangulated graph, such calculations can be performed in time linear in the representation of the clique marginals, i.e. exponential in the size of the cliques. But these are now the cliques of the augmented, triangulated graph. So it is not enough for the Markov network to have small cliques, in order for computations to be tractable, we need the Markov network to have a triangulation with small cliques.

This property is captured by the *tree-width* of a graph:

**Definition 1 (Tree-width).** *The tree-width of a graph is the minimum, over all triangulations of the graph, of the maximum clique size in the triangulation, minus one:*

$$\text{Tree-width}(G) = \min_{\substack{G' \supset G \\ G' \text{ is triangulated}}} \max_{h \in Clique(G)} |h| - 1$$

In this work, we study the problem of projecting a distribution onto Markov networks over graphs of tree-width at most $k$, for some specified $k$. Graphs of tree-width one constitute the class of acyclic graphs (or forests), and so for $k = 1$ this becomes the problem of projecting onto Markov trees. As the width is increased, more complex Markov networks are allowed, with an exponential dependence on $k$.

## 3 Decomposing the Information Divergence

Chow and Liu [CL68] showed that, for Markov trees, the reduction in information divergence, relative to the empty graph, can be additively decomposed to edges. The contribution of each edge of the tree is independent of the structure of the tree, and is the relative information between its nodes. We show a similar decomposition for "wider" triangulated networks. A key point of this decomposition is that the contribution of local elements are *independent* of the graph structure.

Recall that a Markov network can always be factored over its cliques. That is, any distribution $P_X$ that is a Markov network over some graph $G$ can be written as:

$$P_X(x) = \prod_{h \in Clique(G)} \phi_h(x_h) \quad (1)$$

where $\phi_h$ depends only on the outcome $x_h$ inside the clique $h$.

In the general case, the clique factors $\phi_h$ might have a very complex dependence on the distribution. However, when $G$ is triangulated, the factor of a clique depends only on the marginal distribution inside the clique. Moreover, the clique factors can be calculated explicitly and directly from the clique marginals.

We will concentrate on a specific explicit factorization, given by:

$$\phi_h(x_h) = \frac{P_h(x_h)}{\prod_{h' \subset h} \phi_{h'}(x_{h'})} \quad (2)$$

where the product in (1) is taken over all, not necessarily maximal, cliques. That is, we refer here to any complete subgraph of $G$ as a *clique*. Factors corresponding to non-maximal cliques can of course be subsumed into some containing maximal clique factor. However, this leads to clique factors that are dependent on the graph structure. The factors given by (2) are unique in that a clique's factor does not depend on the graph $G$, except for the fact that $G$ includes the clique.[1] A clique's factor depends *only* on the marginal inside the clique, and is completely oblivious to the distribution outside the clique, or even to the structure of the graph outside the clique. This very strong locality property will be essential later.

---

[1] More precisely: consider mappings $P_h \mapsto \phi_h$ from marginal distributions over subsets of variables, to factors over the subset. The mapping given in (2) is the only such mapping, such that (1) holds for *every* triangulated graph $G$ and *every* Markov network $P$ over $G$.

For a specific triangulated graph $G$, the projection of a target distribution $P^{\mathbf{T}}$ onto Markov networks over $G$ can be calculated explicitly. Following (2), and since the projected Markov network is the one in which the clique marginals agree with $P^{\mathbf{T}}$, the projection $\hat{P}_G$ is given by:

$$\hat{P}_G(x) = \prod_{h \in Clique(G)} \hat{\phi}_h(x_h) \qquad (3)$$
$$\hat{\phi}_h(x_h) = \frac{P^{\mathbf{T}}{}_h(x_h)}{\prod_{h' \subset h} \hat{\phi}_{h'}(x_{h'})}$$

where again, the product is over all, not necessarily maximal, cliques.

What we would like to do is to project $P^{\mathbf{T}}$ onto Markov networks over *any* graph of tree-width at most $k$. The core problem is finding the *projected graph* itself (i.e. the graph over which the projected Markov network is achieved). We can associate with every graph $G$ its information divergence from $P^{\mathbf{T}}$—that is, the minimum, over all Markov networks over $G$, of the information divergence from $P^{\mathbf{T}}$:

$$D\left(P^{\mathbf{T}} \| G\right) = \min_{\substack{P \text{ is a} \\ \text{Markov net over } G}} D\left(P^{\mathbf{T}} \| P\right) = D\left(P^{\mathbf{T}} \| \hat{P}_G\right) \qquad (4)$$

The projected graph is the bounded tree-width graph minimizing the information divergence from $G$.

Note that adding edges to a graph can only decrease the information divergence to it, since any distribution that was a Markov network on the sparser graph is certainly also a Markov network on the augmented one. So, the projected graph can always be taken to be triangulated— if it is not triangulated, it can be triangulated by adding edges without increasing the tree-width, yielding an acceptable triangulated graph with lower or equal divergence.

Taking this into account, it is enough to search over all triangulated graphs of tree-width at most $k$ for the graph minimizing the information divergence from $P^{\mathbf{T}}$. But for triangulated graphs, the projected distribution $\hat{P}_G$ is given by (3), and so the information divergence can be calculated as:

$$D\left(P^{\mathbf{T}} \| G\right) = D\left(P^{\mathbf{T}} \| \hat{P}_G\right)$$
$$= \mathbf{E}_{X \sim P^{\mathbf{T}}} \left[\log \frac{P^{\mathbf{T}}(X)}{\prod_{h \in Clique(G)} \hat{\phi}_h(X_h)}\right]$$
$$= \mathbf{E}_{P^{\mathbf{T}}} \left[\log P^{\mathbf{T}}\right] - \sum_{h \in Clique(G)} \mathbf{E}_{X \sim P^{\mathbf{T}}} \left[\log \hat{\phi}_h(X_h)\right]$$

Recall the strong locality of the projected clique factors $\hat{\phi}_h$, i.e. that they depend only on the marginals of $P^{\mathbf{T}}$. Consequentially, for each candidate clique $h$, the term $\mathbf{E}_{P^{\mathbf{T}}} \left[\log \hat{\phi}_h(X_h)\right]$ in the sum depends only on the marginal inside the clique, and not on the structure of the graph $G$.

Consider a weight function over candidate cliques, such that $w(h) = \mathbf{E}_{P^{\mathbf{T}}} \left[\log \hat{\phi}_h(X_h)\right]$. This weight function can be calculated from the target distribution $P^{\mathbf{T}}$ alone, and is independent of the graph. After calculating the weight function for every possible clique, the information divergence to any triangulated graph is given by:

$$D\left(P^{\mathbf{T}} \| G\right) = \mathbf{E}_{P^{\mathbf{T}}} \left[\log P^{\mathbf{T}}\right] - \sum_{h \in Clique(G)} w(h) \qquad (5)$$

where again, the sum is over all, not necessarily maximal, cliques in $G$. In fact, the summation always includes all singleton cliques, i.e. sets of a single vertex. For an empty graph $G$, the only cliques are singleton cliques, and so the information divergence to an empty graph (i.e. fully independent model) is exactly captured by the singleton cliques. Thus, separating out the singletons, we can rewrite (5) as:

$$D\left(P^{\mathbf{T}} \| G\right) = D\left(P^{\mathbf{T}} \| \emptyset\right) - \sum_{h \in Clique(G), |h|>1} w(h) \qquad (6)$$

Equation (6) expresses the reduction in the information divergence versus a simple base model, as a sum of weights (derived from the target distribution) of all non-trivial cliques that appear in the graph. Minimizing the information divergence is thus equivalent to maximizing this sum of weights. This is represented in Figure 1.

Before we return to maximizing the sum of weights, let us investigate the structure of these weights.

## 4 The Weights

The weight of each candidate clique is determined by the target distribution and was defined in terms of the

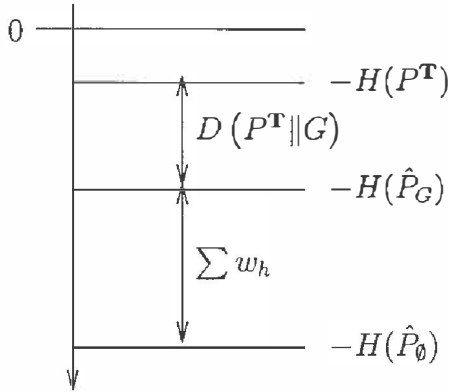

Figure 1: $D\left(P^{\mathbf{T}}\|G\right) = \left(\mathbf{E}_{P^{\mathbf{T}}}[\log \emptyset] - H(P^{\mathbf{T}})\right) - \sum_{h \in Clique(G), |h|>1} w_h$

projected factor:

$$w(h) = \mathbf{E}_{X \sim P^{\mathbf{T}}}\left[\log \hat{\phi}_h(X_h)\right] \quad (7)$$

Incorporating the explicit definition of the projected factors (3):

$$= \mathbf{E}_{P^{\mathbf{T}}}\left[\log \frac{P^{\mathbf{T}}_h(X_h)}{\prod_{h' \subset h} \hat{\phi}_{h'}(X_{h'})}\right]$$

$$= \mathbf{E}_{P^{\mathbf{T}}}\left[\log P^{\mathbf{T}}(X_h)\right] - \sum_{h' \subset h} \mathbf{E}_{P^{\mathbf{T}}}\left[\log \hat{\phi}_{h'} X_{h'}\right]$$

$$= -H(P^{\mathbf{T}}(X_h)) - \sum_{h' \subset h} w(h') \quad (8)$$

where $H(P^{\mathbf{T}}(X_h))$ is the entropy of the marginal distribution of $P^{\mathbf{T}}$ over $h$. This provides for a simple recursive specification of the weights. Unrolling this recursion, the weight of a candidate hyperedge can also be written as a sum:

$$w(h) = -\sum_{h' \subseteq h} (-1)^{|h|-|h'|} H(P^{\mathbf{T}}(X_{h'})) \quad (9)$$

Note that he weight $w(\{v\})$ of a singleton $\{v\}$ is the negative entropy $-H(X_v)$ of the single variable. Accordingly, all singleton weights are negative (or at least non-positive). This is not surprising, as these weights sum up to the negative cross-entropy $\mathbf{E}_{P^{\mathbf{T}}}\left[\log \hat{P}_\emptyset\right]$ and are incorporated in our formulation (6) of the projection problem as part of $D\left(P^{\mathbf{T}}\|\emptyset\right)$. They do not actually appear as weights in the maximum hypertree problem.

However, as more edges are added to the graph, the admissible distributions are less limited and the projected distribution can become more similar to the target distribution. This means that weights of candidate cliques beyond singletons should generally have a positive contribution, representing the reduction in the information divergence, or equivalently the gain in negative cross-entropy.

In fact, the weight of a vertex pair $\{u, v\}$, i.e. an edge, is (following (9)): $w(\{u, v\}) = -H(X_u, X_v) + H(X_u) + H(X_v) = I(X_u; X_v)$, in agreement with Chow and Liu. The mutual information across an edge precisely captures the reduction in information divergence attained by taking into account the dependence between the endpoints.

Now consider the weight of some candidate three-clique $\{1, 2, 3\}$. If the three variables $X_1, X_2, X_3$ are pairwise independent, but have some three-way dependence, then using (9), we can calculate $w(1, 2, 3) = H(X_1) + H(X_2) + H(X_3) - H(X_1, X_2, X_3)$. The weight is again non-negative. It correctly captures the benefit of taking into account the dependency between the variables, as quantified by the reduction in the information divergence. In fact, the weight is equal to the information divergence between the true three-way marginal distribution, and the product distribution of the three singleton marginal distributions.

It is tempting to adopt this clean interpretation of the weights, by which the weight of a candidate clique represents the reduction in information divergence attained by taking into account the additional dependency. Under this interpretation, the weight of a candidate $d$-clique should be the information divergence between the true $d$-way marginal and the maximum entropy $d$-way distribution that agrees with the marginals of all $d-1$ sub-cliques.

The reality is different. Consider a Markov chain over three variables $X_1 \to X_2 \to X_3$. In this case all the information is in the pairwise dependencies, and by the above suggested interpretation, the weight of the candidate three-clique $\{1, 2, 3\}$ should have been zero. Using (9), however, we can check that $w(1, 2, 3) = -I(X_1; X_3) < 0$. On second thought, this should not surprise us. Consider the total weight of a graph containing the three-clique. All the dependencies in the Markov chain are already captured by the two pairwise dependencies $(X_1, X_2)$ and $(X_2, X_3)$. Accordingly, all of the reduction in the information divergence is captured by $w(\{1, 2\}) + w(\{2, 3\})$. However, since we sum all the cliques in the graph, in addition to these two pairwise weights, we will also include $w(\{1, 3\}) = I(X_1; X_3)$. The pairwise weights thus overcount the reduction in the information divergence. The weight of the candidate three-clique, which will always be included if all three pairwise weights are included, accounts for this overcounting.

The weights of candidate cliques thus serve not only to reward for the new dependency made possible by the clique, but also to account for overcounting inside the clique. Pairwise weights need not perform any such accounting, but the weight of any larger candidate clique can be either positive or negative.

The possibility of negative weights is very problematic from an algorithmic standpoint. Negative weights may cause many combinatorial optimization approaches to fail. In fact, the algorithms presented in [KS01] for the maximum weight hypertree problem do not work with arbitrary negative weights. Fortunately, the weights we define here do have some positive structure.

Although for any particular candidate clique of more than two vertices, the clique weight can be negative, it is not possible for too many weights to be negative. As argued before, adding edges to a graph *does* reduce the information divergence, but it can add more then just one additional clique. The total weight of a triangulated graph $G$ is no less than the total weight of a triangulated sub-graph $G'$ of $G$. Accordingly, the sum of the weights of cliques in $G$ that are not cliques in $G'$ must be non-negative.

We call a weight function obeying such a constraint a *monotone* weight function. It is enough to require that the weight function be monotone on cliques, i.e. that the total summed weight of a graph containing a single clique $\{v_1, v_2, \ldots, v_d\}$ is no less than the total summed weight of the graph containing the single clique $\{v_1, v_2, \ldots, v_{d-1}\}$. A weight function that is monotone on cliques is also monotone on all triangulated graphs.

Note that these arguments hold only for triangulated graphs. Otherwise the total summed weight of the graph does not represent any meaningful information quantity as the product of the factors may not be a valid distribution function, let alone the projected distribution.

## 5  The Reduction

The problem of projecting a distribution onto Markov networks of bounded tree-width $k$ can thus be reduced to finding a triangulated graph of bounded tree-width (or equivalently, clique size) that maximizes the total summed weight of its (not only maximal) cliques with respect to some monotone weight function. That is, if we knew an algorithm that finds such a graph, we could use it for projecting distributions onto Markov networks of bounded tree-width.

To do so, we would first calculate a weight for each set of at most $k+1$ vertices (including singletons), starting with the small sets and proceeding to the larger ones, using the recurrence (8). We would then find the maximum weight bounded tree-width triangulated graph for these weights, but ignoring the singleton weights (the singleton weights are necessary only for the recursive calculation).

For tree-width one, i.e. triangulated graphs with no three-cliques, this is the problem of finding a maximum weight tree, and as we saw before, the weights are in agreement with Chow and Liu. For higher tree-width, this is the problem of finding a maximum weight *hypertree* [KS01].

Although the recursive definition provides for a relatively quick method of calculating the weights for small $k$, it is still necessary to calculate $\binom{n}{k+1} = O(n^{k+1})$ weights, taking $O(n^{k+2})$ time.

As we have not considered the representation of the target distribution, we cannot discuss the complexity of the reduction in terms of the problem 'size', as this of course depends on the representation. We do not want to go into the issues of input representations of the distribution, except for one special case which originally motivated us: the case in which the distribution is an empirical distribution of some sample.

The "input representation" in this case is the sample itself, of size $O(Tn \log m)$, where $T$ is the sample size and $m$ is the number of possible outcomes for each random variable. So, if $k$ is part of the input, the reduction is *not* polynomial in the sample, as it is exponential in $k$ while the sample is independent of it. If $k$ is constant, then the reduction is polynomial.

As the number of parameters in the resulting model, and therefore the complexity of calculations on the resulting distribution, is also exponential in $k$, it is tempting to hope that the reduction is comparable to, or at least polynomial in, the resulting number of parameters. This is essentially the output size of the learning problem, and practically also a bound on the input size, as one would generally not have less data then there are parameters to learn. However, this is *not* the case. The number of parameters is only $O(nm^{k+1})$. Therefore if $n \gg m$, the reduction is super-polynomial even in the resulting number of parameters.

## 6  The Reverse Reduction

In order to use the formulation (6) to analyze the computational hardness, we must show how to perform the reverse reduction, i.e. transform an input of the maximum hypertree problem (a weight function) to an input of the projection problem (a distribution) so that the projected distribution implies the maximum hypertree. In this section, we show that for every nonnegative weight function on vertex sets of fixed size, there exists a distribution that yields weights proportional to this set of weights. We thus demonstrate that the problem of finding a maximum hypertree, at least for a non-negative weight function on vertex sets of a fixed size, can be reduced to projecting a distribution onto Markov networks of bounded tree-width.

Furthermore, a "small" sample can be constructed, with an empirical distribution yielding weights that are close enough to these weights, conserving the exact structure of the projected graph. This establishes that the problem of finding a maximum hypertree (for non-negative weights on vertex sets of fixed size) can also be reduced to finding a maximum likelihood Markov network for empirical data.

This reduction is weak, in the sense that the sample size needed to produce specific weights is polynomial in the *value* of the weights (and so exponential in the size of their representation). Still, this pseudo-polynomial reduction is enough to show NP-hardness of finding a maximum likelihood Markov network of bounded tree-width, even for tree-width two.

### 6.1 A distribution yielding desired weights

For a given weight function $w : \binom{V}{k+1} \to [0,1)$ on candidate cliques of size exactly $k+1$, we will consider each vertex as a binary variable and construct a distribution $P_w$ over these variables. The distribution will be such that using it as a target distribution in (7) will yield weights $w'$ proportional to $w$. We will assume, without loss of generality, that $\forall_h w(h) < 1$.

The distribution $P_w$ will be a uniform mixture of $\binom{n}{k+1}$ distributions $P_w^h$, one for each $h \in \binom{V}{k+1}$. Each such $P_w^h$ will deviate from uniformity only by a bias of $r(h)$ in the parity of the variables in $h$. We show below how to select $r(h)$ according to $w(h)$. Explicitly:

$$P_w^h(x) = \begin{cases} \frac{1+r(h)}{2^{|V|}} & \text{If } \sum_{v \in h} x_v \text{ is odd} \\ \frac{1-r(h)}{2^{|V|}} & \text{If } \sum_{v \in h} x_v \text{ is even} \end{cases} \quad (10)$$

This results in a mixed distribution $P_w$ in which all marginals over at most $k$ variables are uniform (and therefore have zero corresponding weight), while the marginal over a set $h$ of exactly $k+1$ variables has a bias of $b = \frac{r(h)}{\binom{n}{k+1}}$. The corresponding weight is therefore

$$w'(h) = -H(X_h) - \sum_{h' \subset h} w(h) = \sum_{v \in h} H(X_v) - H(X_h)$$

$$= |h| \times 1 - H(X_h) = (k+1) + \sum_{x_h} P_w(x_h) \log P_w(x_h)$$

$$= (k+1) + 2^k \frac{1+b}{2^{k+1}} \log \frac{1+b}{2^{k+1}} + 2^k \frac{1-b}{2^{k+1}} \log \frac{1-b}{2^{k+1}}$$

$$= \tfrac{1}{2}\left((1+b)\log(1+b) + (1-b)\log(1-b)\right) \quad (11)$$

Using the natural logarithm and taking the Taylor expansion:

$$= \sum_{i=2 \text{ even}} \frac{b^i}{i(i-1)} = \frac{b^2}{2} + O(b^4) = \frac{r(h)^2}{2\binom{n}{k+1}^2} + O(r(h)^4) \quad (12)$$

Choosing $r(h)$ to be approximately $\sqrt{w(h)}$ (or more precisely, the inverse function of (11)) yields weights proportional to $w$.

### 6.2 A sample yielding desired weights

We have shown a distribution that produces weights proportional to any desired non-negative weight function. But since the biases in this distribution might be irrational (being the inverse of (11)), there is no finite sample that has such a distribution as its empirical distribution.

We will show a finite sample that results in weights that are close enough to the desired weights, such that the optimal structure is conserved. Given a rational weight function $w$, we will show a sample with empirical distribution $\hat{P}_w$ that produces weights $w''(h) = w'(h) + e(h)$ such that $w'$ are proportional to $w$, and $\sum_h |e(h)| < \frac{1}{Q_{w'}}$ where $Q_{w'}$ is the common denominator of $w'$. This is enough, since the total summed $w'$ and $w''$ weights of cliques in the optimal graph will be within $\frac{1}{Q_{w'}}$, less than the possible difference due to taking cliques with differing weights.

We first show how to construct a sample that yields an empirical distribution similar in structure to $P_w$, with rational biases on $k+1$ candidate edges. For any mapping[2] $h \mapsto \frac{p_h}{Q} < 1$ we construct a sample $\mathcal{S}_{\frac{p}{Q}}$ with empirical distribution $\hat{P}_{\frac{p}{Q}}$ such that all $k$-marginals are uniform, and for $|h| = k+1$:

$$\hat{P}_{\frac{p}{Q}}(x_h) = \begin{cases} (1 + \frac{p_h}{Q\binom{n}{k+1}})2^{-|V|} & \text{If } \sum_{v \in h} x_v \text{ is odd} \\ (1 - \frac{p_h}{Q\binom{n}{k+1}})2^{-|V|} & \text{If } \sum_{v \in h} x_v \text{ is even} \end{cases}$$

Unlike the exact $P_w$, parities of larger sets might be very biased. However, these do not effect the resulting weights when searching for width-$k$ Markov networks.

We will build the sample as a pooling of $\binom{n}{k+1}$ equisized samples $\mathcal{S}_{\frac{p}{Q}}^h$, one for each candidate edge of size $k+1$. Each such $\mathcal{S}_{\frac{p}{Q}}^h$ will be constructed from $Q$ equisized blocks of $(k+1)$-wise uniformly independent sample vectors. But for $p$ of these blocks, we will invert the elements of $h$ appropriately so as to set the parity of $x_h^t$

---
[2]The common denominator $Q$ of the biases may be different than the common denominator $Q_{w'}$

to be odd for all sample vectors in the block. Note that this can be done without disrupting the uniformity of any other set of vertices of size at most $k+1$. The resulting $\mathcal{S}_{\frac{p}{Q}}^{h}$ will be uniform on all subsets of size up to $k+1$, except for a bias of $\frac{p(h)}{Q}$ on $h$. Pooling these together yields the desired empirical distribution.

Using [AS91], $(k+1)$-wise independent blocks can be created of size $2n^{k+1}$, yielding a total sample size of $\binom{n}{k+1} Q 2 n^{k+1} = O(Qn^{2k+2})$, where $Q$ is the common denominator of the rational weights.

We now know how to construct a sample with specified *rational* biases. However, the biases corresponding to rational weights are not rational. We first show how to achieve approximate weights with biases that are square roots of rationals, and then describe how these can be approximated with actual rationals.

We saw in (12) that the biases of the mixture components should be approximately the square roots of the desired weights. Using biases $r'(h) = \sqrt{w(h)}$ yields the following weights (where $b' = \frac{r'(h)}{\binom{n}{k+1}} < 1$):

$$w'(h) = \sum_{i=2 \text{ even}} \frac{b'^i}{i(i-1)} < \frac{b'^2}{2} + \sum_{i=4 \text{ even}} \frac{b'^4}{i(i-1)}$$
$$= \frac{b'^2}{2} + \frac{\ln 4 - 1}{2} b'^4 = \frac{1}{2\binom{n}{k+1}^2} w(h) + e(h)$$

Where:

$$\sum_h |e(h)| < \binom{n}{k+1} \frac{\ln 4 - 1}{2} \frac{(\max w)^2}{\binom{n}{k+1}^4} < \frac{0.19}{\binom{n}{k+1}^3} \max w^2$$

Recall that we would like $\sum_h |e(h)| < \frac{1}{Q_{w'}}$. Since the common denominator $Q_{w'}$ scales linearly with the weights, we can achieve this goal by scaling the weights down. But since the weights may not be square rationals, taking their square root might produce irrational weights. This can be overcome in a similar fashion, by using a rational approximation to the square root.

### 6.3 The reduction and hardness

We saw how to reduce the maximum hypertree problem to the maximum likelihood Markov network problem, with the same $k$, and even if the variables are all binary. Note that our reduction is only pseudo-polynomial, as the sample size needed is polynomial in the value of the weights. However, in [Sre00] we show that the maximum hypertree problem is NP-hard, even with zero/one weights:

**Theorem 1 (proved in [Sre00]).**
*The maximum hypertree problem is NP-hard, even for width two, zero/one weights, and weights only on pairs of vertices. Furthermore, it is NP-hard to approximate to within any constant additive offset.*

This is enough to show NP-hardness of the maximum likelihood Markov network problem, even for bound tree-width two.

## 7 Approximation Algorithms

Although the maximum hypertree problem is NP-hard, we present in [KS01] an integer programming based approximation algorithm for it. For any constant $k$, we show a polynomial-time algorithm that finds a triangulated graph of tree-width at most $k$, which has a total summed weight within a constant factor of the maximum possible total summed weight. Unfortunately, this constant approximation factor depends heavily on $k$—for width $k$, we find a graph with total summed weight at least $1/(8^k k!(k+1)!)$ of the optimal. Algorithms with better approximation ratios may be possible, perhaps even with approximation ratios that do not depend on $k$. We discuss how this type of approximation for the combinatorial problem translates into a sub-optimal solution for the maximum likelihood learning problem, as well as the general projection problem.

Recall the decomposition of the information divergence that was presented in Figure 1. When the target distribution is the empirical distribution, the negative cross entropy relative to it is exactly the log likelihood. Figure 1 can be viewed as representing the maximum log likelihood of Markov networks over $\emptyset$ (fully independent models), Markov networks over $G$, and the maximum attainable log likelihood (the negative entropy of the empirical distribution). The weight of the graph is then the gain in maximum log likelihood versus the fully independent model. A constant factor approximation on the weight of the graph translates to a constant factor approximation on the gain in log likelihood.

We can thus attain a constant factor (for constant width) approximation on the gain in log likelihood. But this means we only get a constant *exponential* factor approximation on the likelihood itself. Unfortunately, we cannot hope for much more. Since the maximum hypertree problem is NP-hard to approximate to within any additive constant, we can conclude that it is NP-hard to approximate the likelihood to within any multiplicative constant. That is, for any constants $k > 1$ and $c$, it is NP-hard to find a Markov network of tree-width at most $k$, with likelihood at least $c$ times the optimal likelihood among Markov networks with tree-width at most $k$.

## 8 Discussion

We demonstrated how the problem of projecting a distribution onto Markov networks of bounded tree-width can be cast as a combinatorial optimization problem of finding a maximum weight hypertree. By studying the maximum hypertree problem, we were able to prove that the projection problem is NP-hard and to provide an approximation algorithm with a provable performance guarantee. The approximation ratio is rather weak, and a large gap remains between this positive result and the computational hardness result. However, now that the maximum weight hypertree problem has been presented to the algorithms community, further progress on it will directly correspond to improved algorithms for projecting a distribution onto Markov networks of bounded tree-width.

In Section 5 we discussed how a constant factor approximation for maximum weight hypertree translates to a constant factor approximation on the reduction in information divergence (or, equivalently, the gain in likelihood). As can be seen in Figure 1, a constant factor approximation of the weight of the graph does *not* provide for a constant factor approximation of the information divergence itself. If the target distribution is far from being a narrow Markov network, or has low entropy (as is usually the case for an empirical distribution), than approximating the reduction in information divergence is a more stringent, more useful requirement than approximating the information divergence itself. In fact, approximating the gain in maximum likelihood is always more stringent then approximating the maximum likelihood itself. However, approximating the information divergence may be interesting when the target distribution is "almost" a narrow Markov network. In this case, approaching the optimal information divergence to within a small factor is much more stringent then approximating the reduction versus an independent model. This might be relevant even if the target distribution is an empirical distribution, e.g. if it is a large empirical sample from a narrow Markov network.

It is also interesting to study the weights that carry the decomposition of the reduction in information divergence. In particular, we might ask if the monotonicity is the true property defining the structure of these weights. That is, is any monotone weight function realizable by some distribution? This question could potentially be answered by extending the reverse reduction of Section 6 from a positive weight function on sets of a fixed number of vertices, to any monotone weight function.

In this work we concentrated on finding maximum likelihood models. It would be interesting to extend this work also to scoring functions that are appropriate for model selection. In fact, minimum description length (MDL) scores can be decomposed to clique weights over triangulated graphs. However, the weights are no longer monotone and the approximation results do not hold. Moreover, although the optimal MDL score might be achieved on a non-triangulated graph, the weights sum up correctly only on triangulated graphs. The hardness results do carry over to MDL scores, i.e. finding the *triangulated* graph of bounded tree-width that minimizes its MDL is NP-hard.

### Acknowledgements

I would like to thank Tommi Jaakkola for many helpful discussions and help in preparing this paper, and David Karger for working with me on the approximation algorithm presented in the companion paper [KS01].